%% file: emnlp2022.tex
\pdfoutput=1

\documentclass[11pt]{article}
\usepackage{authblk}

\usepackage{EMNLP2022}

\usepackage{times}
\usepackage{latexsym}
\usepackage{tabu}
\usepackage{multirow}
\usepackage{lipsum,multicol}
\usepackage{hhline}
\usepackage[pdftex]{graphicx}

\usepackage[T1]{fontenc}

\usepackage[utf8]{inputenc}

\usepackage{microtype}

\usepackage{inconsolata}

\title{\textsc{Pneg}: Prompt-based Negative Response Generation\\for Dialogue Response Selection Task}
\author{\textbf{Nyoungwoo Lee$^1$}}
\author{\textbf{ChaeHun Park$^2$}}
\author{\textbf{Ho-Jin Choi$^3$}}
\author{\textbf{Jaegul Choo$^2$}}
\affil{$^1$Scatter Lab \hspace{0.3cm} $^2$KAIST AI \hspace{0.3cm} $^3$KAIST}
\affil{\texttt{\small\ nyoungwoo@scatterlab.co.kr, \{ddehun, hojinc, jchoo\}@kaist.ac.kr}}

\begin{document}
\maketitle
\begin{abstract}
In retrieval-based dialogue systems, a response selection model acts as a ranker to select the most appropriate response among several candidates.
However, such selection models tend to rely on context-response content similarity, which makes models vulnerable to adversarial responses that are semantically similar but not relevant to the dialogue context. 
Recent studies have shown that leveraging these adversarial responses as negative training samples is useful for improving the discriminating power of the selection model.
Nevertheless, collecting human-written adversarial responses is expensive, and existing synthesizing methods often have limited scalability.
To overcome these limitations, this paper proposes a simple but efficient method for generating adversarial negative responses leveraging a large-scale language model.
Experimental results on dialogue selection tasks show that our method outperforms other methods of synthesizing adversarial negative responses. 
These results suggest that our method can be an effective alternative to human annotators in generating adversarial responses. Our dataset and generation code is available at \url{https://github.com/leenw23/generating-negatives-by-gpt3}.
\end{abstract}

\input{EMNLP 2022/1.Introduction}
\input{EMNLP 2022/2.Method}
\input{EMNLP 2022/3.Experimental_Setup}
\input{EMNLP 2022/4.Experiments}
\input{EMNLP 2022/5.Conclusion}
\input{EMNLP 2022/6.Limitations}
\input{EMNLP 2022/7.Ethics_Statement}

\section*{Acknowledgements}
This work was supported by Institute for Information \& communications Technology Planning \& Evaluation(IITP) grant funded by the Korea government(MSIT) (No. 2020-0-00368, A Neural-Symbolic Model for Knowledge Acquisition and Inference Techniques, No. 2013-2-00131, Development of Knowledge Evolutionary WiseQA Platform Technology for Human Knowledge Augmented Services, and No. 2019-0-00075, Artificial Intelligence Graduate School Program(KAIST))

\bibliography{anthology,emnlp2022}
\bibliographystyle{acl_natbib}

\appendix
\clearpage
\input{EMNLP 2022/appendix/1.Appendix_prompt_design}
\clearpage
\input{EMNLP 2022/appendix/2.Appendix_other_experiments}
\clearpage
\input{EMNLP 2022/appendix/3.Appendix_prompt}
\clearpage
\input{EMNLP 2022/appendix/4.Appendix_main_examples}
\clearpage
\input{EMNLP 2022/appendix/5.Appendix_ablation_examples}

\end{document}

%% file: EMNLP 2022/1.Introduction.tex
\section{Introduction}
\label{introduction}

In retrieval-based dialogue systems, the response selection model aims to predict the most appropriate response among multiple candidates retrieved for a given conversation context~\citep{zhou-etal-2018-multi, wu2019sequential, chen-etal-2022-contextual}. The selection model is generally trained to distinguish a related (positive) response from randomly sampled negative responses on training datasets, but such a model generally poses the following problems.

\begin{figure}[t]
\centering
\begin{tabular}{c}
     \includegraphics[width=0.48\textwidth]{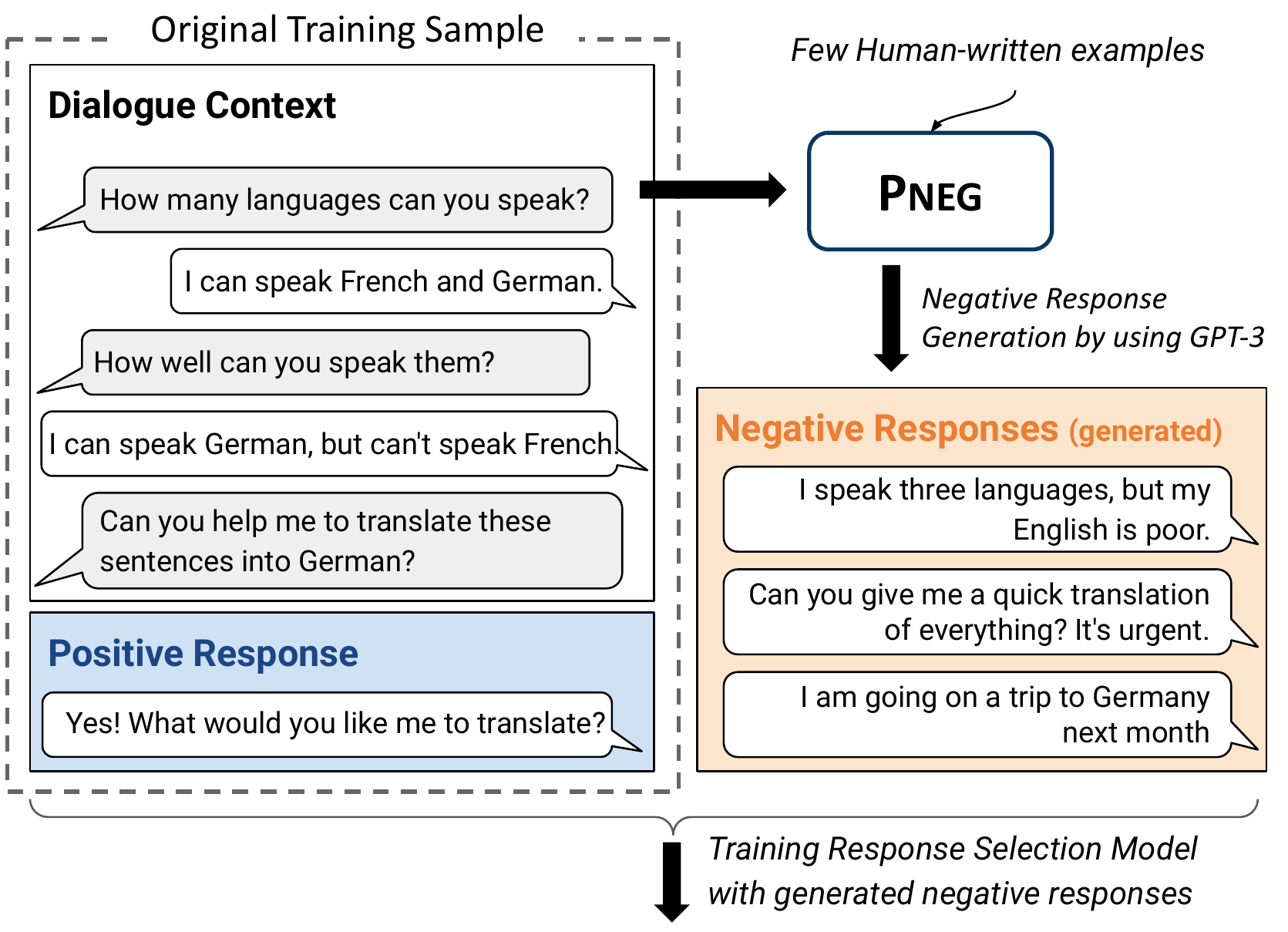}
\end{tabular}
\caption{A conceptual pipeline of prompt-based negative response generation.}
\label{fig:method}
\vspace{-0.1in}
\end{figure}

First, randomly selected negatives are often too easy to distinguish because they are totally irrelevant to the dialogue context~\citep{li-etal-2019-sampling, lin-etal-2020-world}. In this case, the model is more likely to predict the response only by relying on the superficial content similarity of the context-response pairs~\citep{yuan-etal-2019-multi, sai-etal-2020-improving, whang2021response}. These models are vulnerable to adversarial responses with high content similarity to the dialogue context, and fail to distinguish subtle differences in various contexts in real-world scenarios. Second, random sampling causes sampling bias, such as containing false negatives for a given dialogue context~\citep{zhou-etal-2022-debiased}. These biases inherent in the training datasets hinder the accurate prediction of the selection model, resulting in performance degradation.

To mitigate this problem, recent studies have proposed various methods to synthesize and leverage adversarial negative training samples so that the selection model can learn features beyond content similarity~\citep{srivastava2020robustness, kaushik2021learning}. However, existing methods for synthesizing adversarial negative responses~\citep{ebrahimi-etal-2018-hotflip, alzantot-etal-2018-generating, zhang-etal-2019-generating-fluent, qiu2021challenging, gupta-etal-2021-synthesizing} still have limitations in creating human-like responses. The most reliable method is to collect human-written adversarial negatives~\citep{sai-etal-2020-improving}, but it is not scalable because it is expensive and time consuming.

To overcome these limitations, we note that large-scale language models such as GPT3 can be utilized as a low-cost data labeler~\citep{wang-etal-2021-want-reduce}. In this paper, we present \textsc{Pneg}, a \textbf{P}rompt-based \textbf{NE}gative response \textbf{G}eneration method leveraging a large-scale language model (Figure~\ref{fig:method}). In P-NEG, we can cheaply collect human-like adversarial negatives by using an in-context learning-based data augmentation method with human-written samples as demonstration examples.

Experimental results on the dialogue response selection task show that the selection model that trained the negative samples from \textsc{Pneg} has better performance than other baselines. We then conduct quality evaluation and ablation studies to analyze the validity of \textsc{Pneg}. Consequently, we argue that \textsc{Pneg} can be an efficient alternative to human annotators in generating adversarial responses.

Our contributions are as follows:
\begin{itemize}
    \item We propose \textsc{Pneg}, a \textbf{P}rompt-based \textbf{NE}gative response \textbf{G}eneration method. 
    \item Our method can generate high-quality adversarial negative responses only with a few human-written examples.
    \item We show that our method outperforms other baselines across multiple model architectures on the response selection task.
\end{itemize}

%% file: EMNLP 2022/2.Method.tex
\section{\textsc{Pneg}: Prompt-based NEgative response Generation}
\label{promptconstruction}
Large-scale language models such as GPT-3~\citep{NEURIPS2020_1457c0d6} can augment fluent text training samples using natural language prompts and in-context examples~\citep{yoo2021gpt3mix, schick-schutze-2021-generating, bae2022building, liu2022wanli}. By extending such direction to the response selection task, we propose \textsc{Pneg}, a \textbf{P}rompt-based \textbf{NE}gative response \textbf{G}eneration method for robust response selection models.
Our method consists of three steps: (1) selecting demonstration examples, (2) constructing a prompt containing examples and target dialogue context, and (3) generating adversarial negative responses by inputting the prompt into GPT-3. The generated negative responses are used as training samples for response selection task.

\begin{figure}[t]
\centering
\begin{tabular}{c}
     \includegraphics[width=0.48\textwidth]{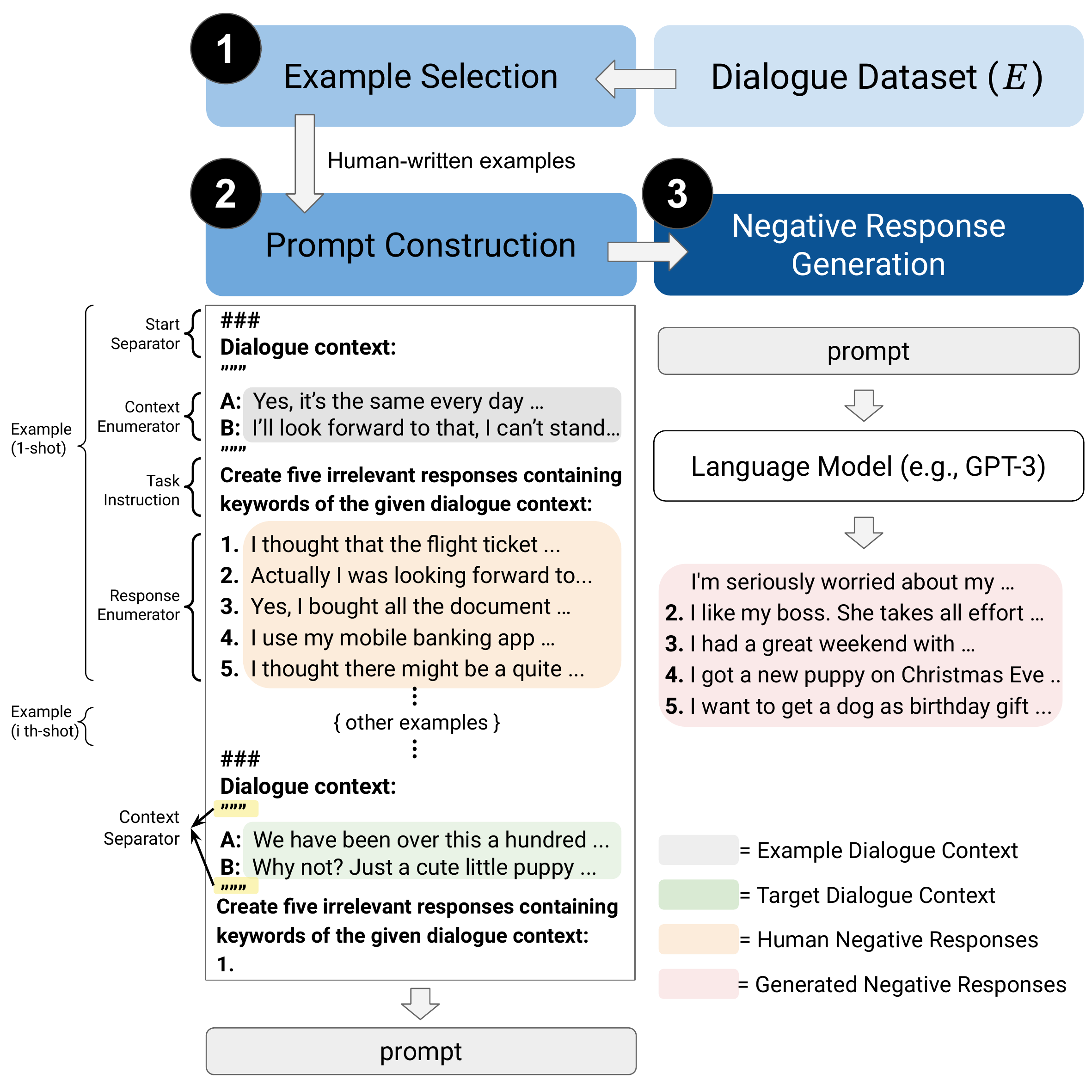}
\end{tabular}
\caption{Overall pipeline of \textsc{Pneg}.}
\label{fig:promptconfig}
\vspace{-0.1in}
\end{figure}

\subsection{Step 1: Example Selection}
We first sample a total $k$ demonstration examples from an example set $E$ to construct a prompt for in-context learning of GPT-3~(A in Figure~\ref{fig:promptconfig}). We consider \textit{DailyDialog++}~\cite{sai-etal-2020-improving} as an example set to generate training samples that have similar patterns as human-written high-quality linguistic patterns. The example set consists of a dialogue context, a positive response, and multiple human-written negative responses. We uniformly sampled examples from the dialogue dataset. The context and human-written negative responses are used in the following prompt construction step.

\subsection{Step 2: Prompt Construction}
\paragraph{Prompt Design}
We first design a prompt to use as an input to GPT-3. Large-scale language models are generally pre-trained to generate responses with high coherence for a given context, but our method aims to generate adversarial responses with high content similarity but low coherence. In particular, natural language prompt-based reasoning is sensitive to prompt changes~\citep{jin-etal-2022-good}, so we have to carefully design prompts to generate accurate negative responses. We devise several prompts to evaluate generative quality, inspired by related works~\citep{yoo2021gpt3mix, schick-schutze-2021-generating, min2022rethinking}. We evaluate the devised prompts by the response selection task, and then select the finest prompt template based on the results of Appendix~\ref{sec:appendix_prompt_changes}.

\paragraph{Prompt Construction}
Our prompt is constructed with the chosen prompt template, $k$ number of examples, and the target dialogue context that we aim to generate multiple negative responses. The prompt template consists of three components to clarify the role of each example and target context: (1) a task instruction $I$ written in natural language, (2) an enumerator to receive each utterance from examples and the target context, and (3) a separator to separate each example or dialogue context in the prompt. The details of each component in the prompt template are shown in Figure~\ref{fig:promptconfig}.

\subsection{Step 3: Negative Response Generation}
Now, GPT-3 can generates augmented negative responses following our input prompt~(C in Figure~\ref{fig:promptconfig}). The examples within the prompt encourage the language model to generate negative responses of similar patterns to the human-written negative responses. The task instruction directly guides the model to understand the target task and the relationship between a dialogue context and corresponding negative responses in the examples.

%% file: EMNLP 2022/3.Experimental_Setup.tex
\section{Experimental Setup}
\label{experimental_setup}
\subsection{Dialogue Response Selection Task}
We evaluate our method on the dialogue response selection task. We train the selection model with 11 response candidates consisting of 1 positive response, 5 random and 5 adversarial negative responses per context. We report the R@1 and mean reciprocal rank~(MRR) score. For evaluation, we use random and adversarial test datasets which consist of the 6 candidates for each context.

\subsection{Datasets}
\paragraph{DailyDialog++}
We use the \textit{DailyDialog++}~\cite{sai-etal-2020-improving} dataset for our overall experiments. This dataset consists of 16,900, 1028, and 1142 dialogue contexts in training, validation, and test datasets, respectively. Since only the subset of 9259 contexts in the training dataset contains adversarial responses, we use them as our training dataset. Each context has five adversarially curated negative responses written by human annotators.

\paragraph{PersonaChat}
We also use the \textit{PersonaChat} dataset~\cite{zhang2018personalizing} on the response selection task. This dataset consists of 8938, 1000, and 968 dialogue conversations in training, validation, and test datasets, respectively. We use 8938 contexts for training, and concatenate the persona sentences in front of the context. Since there are no human-written adversarial negative responses in this dataset, we create an adversarial test dataset by sampling one response from the context and including it in the candidate responses following \citet{gupta-etal-2021-synthesizing} and \citet{whang2021response}.

\subsection{Baselines}
\label{baselines}
We compare our method with following baselines.
\paragraph{Random} Randomly sampled responses.
\paragraph{Human~\citep{sai-etal-2020-improving}} Human-written adversarial responses in \textit{DailyDialog++} dataset.
\paragraph{BM25~\citep{karpukhin-etal-2020-dense}} Retrieved responses from BM25~\cite{bm25}.
\paragraph{Semi-hard~\citep{li-etal-2019-sampling}} Retrieved responses from training dataset based on their similarity between positive response with a margin of $\alpha$. We perform a static sampling using sentence-BERT~\cite{reimers-gurevych-2019-sentence} with $\alpha$ as 0.07 following~\citet{gupta-etal-2021-synthesizing}. 
\paragraph{Mask-and-fill~\citep{gupta-etal-2021-synthesizing}} This method first randomly masks the words in a answer response, and infill them using masked language modeling conditioned on a random context.
\paragraph{Key-sem~\citep{gupta-etal-2021-synthesizing}} This method generates new responses conditioned on keywords in the context using GPT-2~\cite{gpt2}.
\paragraph{\textsc{Pneg} (Ours)} GPT-3 generated adversarial negative responses by using our method, \textsc{Pneg}.

\subsection{Models}
\label{Models}
We train dialogue selection models with different negative responses described in \S\ref{baselines}. The models are based on cross-encoder architecture, and three different pre-trained language models are used in experiments: 1) BERT~\cite{devlin2019bert}, 2) RoBERTa~\cite{liu2019roberta}, and 3) ELECTRA~\cite{clark2019electra}. The implementation details are provided in Appendix~\ref{sec:appendix_implementation}.

%% file: EMNLP 2022/4.Experiments.tex
\section{Experiments}
\label{experiments}
\subsection{Performance on Response Selection Task}
\label{exp:response_selection}
We compare our method with the baselines for the response selection task\footnote{The negative responses samples generated by each method are provided in Table~\ref{table:response_examples}.} (Table~\ref{table:exp1}). On the adversarial test dataset, we notice that \textsc{Pneg} consistently outperforms other baselines across different model architectures. The \textsc{Pneg} especially shows the most similar performance to the human baseline, which suggests that our method can be an effective alternative to human annotators for collecting adversarial negative samples.
In the random test dataset, Semi-hard performs better than other baselines including \textsc{Pneg} and even the human baseline. This tendency indicates that the robustness of the models to the adversarial test dataset does not always lead to the random test dataset.
We speculate that these results are due to data distribution shifts according to different negative response sampling strategies~\citep{penha2021calibration}.
Retrieval-based methods like BM25 and Semi-hard would suffer from finding appropriate negative responses when the dialogue corpus for response retrieval is limited.

We also compare our method with the baselines for the response selection task on the \textit{PersonaChat} (Table~\ref{table:persona}). Although \textsc{Pneg} generates negative responses using human-written examples from \textit{DailyDialog++}, it shows better performance than other baselines in the adversarial test dataset. Such results prove the scalability of \textsc{Pneg} across multiple dialogue datasets, because our method can automatically generate adversarial negative samples with only a few human-written samples. It means that our method can easily transfer to other domain conversations only with minimal human effect. Since we used fewer samples for \textit{PersonaChat} experiments, the response selection models overall show lower performance than the \textit{DailyDialog++} experiment.

\input{0.Main-table}

\input{0.Sub-table}

\subsection{Synthetic Dataset Quality}
\label{synthetic_quality}
\begin{figure}[t]
\centering
\begin{tabular}{l}
     \includegraphics[width=0.46\textwidth]{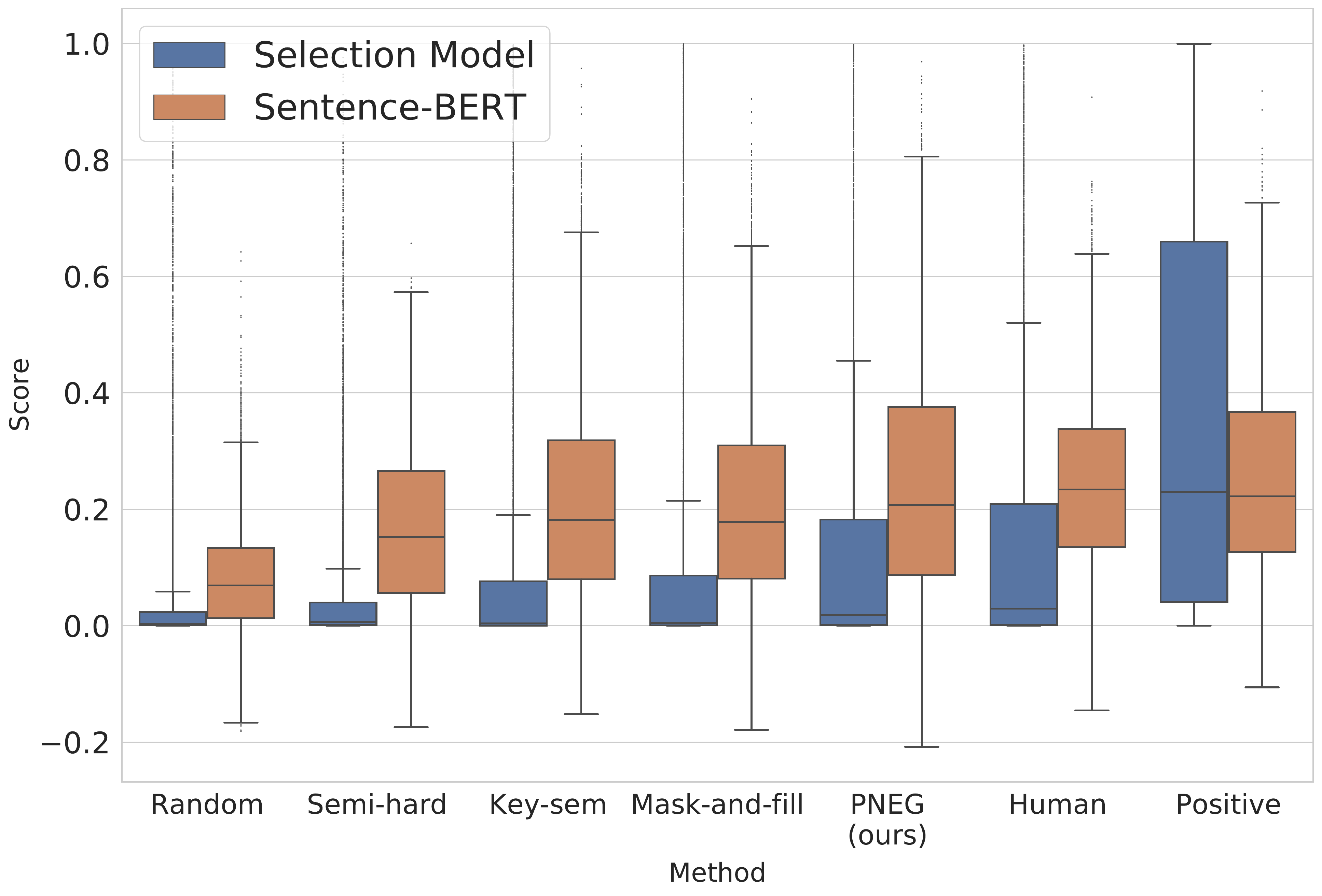}
\end{tabular}
\caption{Box plot of prediction scores (blue) and similarity score (orange) for each type of response. The prediction scores are linearly normalized into the [0,1].}
\label{fig:box_plot}
\end{figure}

We evaluate the quality of adversarial responses with predictive scores of the response selection model and context-response similarity model\footnote{The experiment details are provided in Appendix~\ref{sec:appendix_auto_eval}.}. We assume that the higher the prediction score of the selection model for the adversarial negative response, the more effective it is for the robust training of the model. The evaluation results are shown in Figure~\ref{fig:box_plot}. In both models, the prediction score for negative responses generated by \textsc{Pneg} is higher on average than other responses. The difference is more significant in the score of the selection model, suggesting that \textsc{Pneg} can produce more effective adversarial responses that are confused with the positive response. Although Semi-hard samples negative responses using similarity scores from Sentence-BERT, the responses have lower scores than other methods because the sampling pool is limited. We additionally provide human evaluation results for synthetic quality in Appendix~\ref{exp:human_eval}.

\subsection{Varying the Size of Example Set}
\label{exp:ablation}
\begin{table}[t]
\centering
    {\small
    {\tabulinesep=0.6mm
    \begin{tabu}{l|c|c|c|c}
    \textbf{Sub. (\%)} & \textbf{$|E|$} & \multicolumn{2}{c|}{\textbf{Test Set (R@1)}} & \textbf{Mean}\\
    \hhline{~|~|--|~|}
    &  & Random & Adv & \scriptsize{Rand + Adv}\\
    \hline
    0.1 +\textsc{reuse} & 9$+\alpha$ & 0.852 & 0.899 & 0.876\\
    0.1 & 9 & 0.843 & 0.938 & 0.891\\
    1 & 93 & 0.845 & 0.936 & 0.891\\
    10 & 926 & 0.852 & 0.936 & 0.894\\
    100 (\textsc{Pneg}) & 9259 & 0.877 & 0.941 & 0.909\\
    \hline
    \end{tabu}
    }}
\caption{\label{table:ex_set} Performance on the size changes of $E$ containing examples used to construct prompts of our method.}
\vspace{-0.1in}
\end{table}
We study the effect of the size of the example set $E$ containing examples for prompt construction on the performance of the response selection task. As shown in Table~\ref{table:ex_set}, even if the size of $E$ becomes extremely small (e.g., 0.1\%), the performance on the adversarial test dataset hardly decreases. These results show that our method can efficiently generate high-quality adversarial responses by collecting only a few demo examples. To increase the diversity of examples, we further try \textsc{+reuse}, which continuously adds the negative responses generated by \textsc{Pneg} to $E$. However, the 0.1\%\textsc{+reuse} has a significant performance drop in the adversarial test dataset. These results shows that example quality is more important than the diversity of examples to optimize the quality of the generated negative responses.

%% file: 0.Main-table.tex
\begin{table}[t]
\centering
    {\small
    {\tabulinesep=0.5mm
    \begin{tabu}{l|l|cc|cc}
    \textbf{Model} & \textbf{Method} & \multicolumn{4}{c}{\textbf{Test Dataset}}\\
    \hhline{~~|----}
    & & \multicolumn{2}{c|}{Random} & \multicolumn{2}{c}{Adversarial}\\
    \hhline{~~|----}
    & & R@1 & MRR & R@1 & MRR\\
    \tabucline[1pt]{-}
    \scriptsize{BERT} & \scriptsize{Random} & 0.865 & 0.923 & 0.674 & 0.806\\
    & \scriptsize{BM25} & 0.845 & 0.911 & \underline{0.857} & 0.915\\
    & \scriptsize{Semi-hard} & \textbf{0.881} & \textbf{0.934} & 0.672 & 0.804\\
    & \scriptsize{Key-sem} & 0.864 & 0.923 & 0.842 & 0.909\\
    & \scriptsize{Mask\&fill} & \underline{0.869} & \underline{0.926} & 0.856 & \underline{0.916}\\
    & \scriptsize{\textsc{Pneg}} & 0.867 & 0.924 & \textbf{0.937} & \textbf{0.964}\\
    \hline
    & \scriptsize{Human} & 0.870 & 0.926 & 0.954 & 0.974\\
    \tabucline[1pt]{-}
    \scriptsize{RoBERTa} & \scriptsize{Random} & 0.879 & 0.932 & 0.658 & 0.797\\
    & \scriptsize{BM25} & 0.879 & 0.932 & 0.865 & 0.920\\
    & \scriptsize{Semi-hard} & \textbf{0.892} & \textbf{0.937} & 0.660 & 0.797\\
    & \scriptsize{Key-sem} & \underline{0.889} & \textbf{0.937} & \underline{0.868} & \underline{0.924}\\
    & \scriptsize{Mask\&fill} & 0.873 & 0.927 & \underline{0.868} & 0.922\\
    & \scriptsize{\textsc{Pneg}} & 0.882 & \underline{0.933} & \textbf{0.942} & \textbf{0.967}\\
    \hline
    & \scriptsize{Human} & 0.891 & 0.938 & 0.955 & 0.975\\
    \tabucline[1pt]{-}
    \scriptsize{ELECTRA} & \scriptsize{Random} & 0.893 & \underline{0.941} & 0.705 & 0.823\\
    & \scriptsize{BM25} & 0.853 & 0.916 & \underline{0.900} & \underline{0.940}\\
    & \scriptsize{Semi-hard} & \textbf{0.908} & \textbf{0.949} & 0.730 & 0.840\\
    & \scriptsize{Key-sem} & \underline{0.895} & 0.940 & 0.869 & 0.929\\
    & \scriptsize{Mask\&fill} & \underline{0.895} & \underline{0.941} & 0.877 & 0.923\\
    & \scriptsize{\textsc{Pneg}} & 0.873 & 0.928 & \textbf{0.951} & \textbf{0.972}\\
    \hline
    & \scriptsize{Human} & 0.896 & 0.941 & 0.967 & 0.982\\
    \tabucline[1pt]{-}
    \end{tabu}}}
    \caption{\label{table:exp1} Performance in the dialogue response selection task on Random and Adversarial test datasets based on the \textit{DailyDialog++}. Among the methods except for human baseline, the best result is shown in \textbf{bold}, and the second-highest result is \underline{underlined}.}
\end{table}

%% file: 0.Sub-table.tex
\begin{table}[t]
\centering
    {\small
    {\tabulinesep=0.6mm
    \begin{tabu}{l|c|c|c}
    \textbf{Approach} & \multicolumn{2}{c|}{\textbf{Test Set}} & \textbf{Mean}\\
    \hhline{~|--|~|}
    & Random & Adversarial & Rand + Adv.\\
    \hline
    Random & \textbf{0.815} & 0.316 & 0.566\\
    Semi-hard & \underline{0.814} & 0.338 & 0.576\\
    BM25 & 0.718 & \underline{0.637} & \underline{0.678}\\
    \textsc{Pneg} (Ours) & 0.774 & \textbf{0.684} & \textbf{0.729}\\
    \hline
    \end{tabu}
    }}
    \caption{\label{table:persona} Performance in the dialogue response selection task on Random and Adversarial test datasets based on the \textit{PersonaChat}. Among the methods except for human baseline, the best result is shown in \textbf{bold}, and the second-highest result is \underline{underlined}.}
\vspace{-0.1in}
\end{table}

%% file: EMNLP 2022/5.Conclusion.tex
\section{Conclusion}
\label{conclusion}
In this paper, we present \textsc{Pneg}, a prompt-based adversarial negative response generation method for response selection task. \textsc{Pneg} can collect high-quality adversarial negative responses at a lower cost than human annotators. Our experiments on dialogue response selection tasks show that negative responses generated by \textsc{Pneg} can improve the discriminating power of the selection models. In future work, we are interested in designing more confused adversarial responses and how to effectively utilize them for training the selection models.

%% file: EMNLP 2022/6.Limitations.tex
\section*{Limitations}
\label{limitations}
In this section, we present two major limitations of our work.
First, \textsc{Pneg} has not yet been tested on a larger dataset. We expect our method can produce effective adversarial training samples in large-scale datasets at a reasonable cost. However, since the performance of sampling-based methods can improve on sufficiently large training datasets, it is necessary to compare such methods with \textsc{Pneg}.
Second, there is a limitation to leveraging the adversarial negative response for training the selection model in the dialogue response selection task. The selection model that learns the adversarial negative response improves performance in the adversarial test dataset with the same distribution of candidates, but decreases performance in the random test dataset. This is not the result we were hoping for, and we need a close analysis to solve this phenomenon.

%% file: EMNLP 2022/7.Ethics_Statement.tex
\section*{Ethics Statement}
\label{ethics statement}
We clarified compliance with the \href{https://www.aclweb.org/portal/content/acl-code-ethics}{ACL Ethics Policy} in this study. In particular, in the human evaluation experiment, we recruited participants fairly, designed experiments that require minimal effort, and provided adequate rewards to participants.

%% file: EMNLP 2022/appendix/1.Appendix_prompt_design.tex
\twocolumn
\section{Implementation Details}
\label{sec:appendix_implementation}
The inference on GPT-3 was carried out via the Open AI API Beta Access program. We used the largest GPT-3 model, \textit{davinci}. It takes an average of \$0.03 for generating negative responses for one sample of the dialogue dataset, which is approximately one-tenth of the cost of collecting through humans. Each inference consumes an average of 600 tokens and takes an average of 4.28 seconds. For the balance between diversity and quality of synthetic samples from our method, \textsc{Pneg}, we set the temperature to 0.8 and both frequency penalty and presence penalty to 0.4. 

For training of selection models, we predict the score of each context-response pair for every responses in a candidate responses and use cross entropy loss to maximize the score of the context-positive response pair. We use the pre-trained language models\footnote{\url{bert-base-uncased}, \url{roberta-base} and \url{google/electra-base-discriminator} are used.} released by \citet{wolf2019transfertransfo} for experiments. We use the Adam optimizer~\cite{adam} with an initial learning rate as 2e-5, and set the maximum epochs to 3.  We use the validation loss to select the best model. The random seed is fixed, and the batch size is set to 16 per GPU on machines with 2 Nvidia TITAN RTX GPUs. We repeated the experiments three times with different random seeds and report the average performance.

\section{Performance on Prompt Changes}
\label{sec:appendix_prompt_changes}
In this section, we study the performance according to the prompt changes of \textsc{Pneg} on the response selection task.

\subsection{Number of Examples ($k$)}
\label{exp:ablation_k}
\begin{table}[t]
\centering
    {\small
    {\tabulinesep=0.6mm
    \begin{tabu}{l|c|c|c}
    \textbf{k} & \multicolumn{2}{c|}{\textbf{Test Set}} & \textbf{Mean}\\
    \hhline{~|--|~|}
    & Random & Adv & Rand + Adv\\
    \hline
    0 & 0.799 & 0.841 & 0.820\\
    1 & 0.856 & 0.893 & 0.875\\
    2 (\textsc{Pneg}) & 0.859 & 0.928 & 0.894\\
    \hline
    \end{tabu}
    }}
\caption{\label{table:ab_num} Ablation study on the number of examples $k$ in the prompts of our method. ($k$ = 0, 1, and 2)}
\end{table}

\begin{table}[t]
\centering
    {\small
    {\tabulinesep=0.6mm
    \begin{tabu}{l|c|c|c|c}
    \textbf{k} & \textbf{position} & \textbf{Jaccard} & \multicolumn{2}{c}{\textbf{Length Correlation}}\\
    \hhline{~|~|~|--}
    & (pos/k) &\textbf{Similarity} & Pearson & Spearman\\
    \hline
    1 & 1/1 & \textbf{0.046} & \textbf{0.376} & \textbf{0.351}\\
    2 & 1/2 & 0.031 & 0.154 & 0.174\\
    2 & 2/2 & 0.035 & 0.339 & 0.293\\
    2 & all & 0.041 & 0.342 & 0.324\\
    \hline
    \end{tabu}
    }}
\caption{\label{table:ab_cont}  Correlation of generated negative responses in our method with the few-shot examples ($k$>0). We measure the Jacquard similarity and length correlation between the example and the generated response.}
\end{table}

We analyze the effect of the number of examples $k$ in the prompts of our method on the response selection model. The results are in Table~\ref{table:ab_num}. Our method has the highest performance when using two examples ($k$=2), but using one example ($k$=1) also can be a reasonable alternative. The performance of prompts without examples ($k$=0) is rapidly degraded due to frequent occurrence or false-negative generation. These results show that it is important to provide an adequate number of examples to minimize the occurrence of false-negative responses.

We also measured the Jaccard similarity and length correlation between generated responses and each example in the prompt to qualitatively analyze the effect of the example on the generated responses. As shown in Table~\ref{table:ab_cont}, the Jaccard similarity and length correlation coefficient are measured higher when $k$=1 than when $k$=2, and the generated responses are more affected by the closer example. Such contamination effect can increase the effectiveness of the in-context example as guidance of the task, but it can also limit the diversity.

\subsection{Task Instruction Type ($I$)}
\label{exp:task_ins}
\begin{table}[t]
\centering
    {\small
    {\tabulinesep=0.6mm
    \begin{tabu}{l|c|c|c|c}
    \textbf{Type} & \textbf{k} & \multicolumn{2}{c|}{\textbf{Test Set}} & \textbf{Mean}\\
    \hhline{~~|--|~|}
    && Random & Adv & Rand + Adv\\
    \hline
    I\_{dir} & 2 & 0.877 & 0.941 & 0.909\\
    I\_{pos} & 2 & 0.857 & 0.940 & 0.898\\
    I\_{imp} & 0 & 0.788 & 0.800 & 0.796\\
    \hline
    \end{tabu}
    }}
\caption{\label{table:TI} Ablation studies on task instruction changes in the prompt of \textsc{Pneg}. The $I_{pos}$ and the $I_{pneg}$ are follows 2-shot setting, and the $I_{imp}$ follows zero-shot setting.}
\end{table}

We compare the performance of \textsc{Pneg} according to changes in the task instruction. We design the following three types: (1) direct instruction ($I_{dir}$), (2) direct instruction with a positive response ($I_{pos}$), and (3) implicit instruction ($I_{imp}$). We expect that $I_{pos}$ can generate more challenging negatives by referring to the positive response, and $I_{imp}$ can generate diverse responses due to the reduced constraints in the prompt. As shown in Table~\ref{table:TI}, $I_{pos}$ show lower performance than $I_{dir}$ in the random test dataset. Since $I_{imp}$ is vulnerable to false-negative generation, it has the lowest performance in both random and adversarial test datasets.

%% file: EMNLP 2022/appendix/2.Appendix_other_experiments.tex
\twocolumn
\begin{table}[t!]
\centering
    {\small
    {\tabulinesep=0.6mm
    \begin{tabu}{lcc}
    \textbf{Approach} & \textbf{Pred. Score} & \textbf{Similarity}\\
    \hline
    Random & $-2.749_{2.48}$ & $0.078_{0.09}$\\
    Semi-hard & $-2.051_{2.83}$ & $0.161_{0.13}$ \\
    Mask-and-fill & $-1.925_{3.26}$ & $0.207_{0.17}$\\
    Key-sem & $-1.956_{3.34}$ & $0.212_{0.17}$\\
    \textsc{Pneg} (Ours) & $-0.598_{3.53}$ & $0.241_{0.20}$\\
    Human & $-0.279_{3.13}$ & $0.242_{0.15}$\\
    \hline
    Positive & $2.779_{2.25}$ & $0.256_{0.17}$ \\
    \hline
    \end{tabu}
    }}
\caption{\label{table:auto_quality_eval} Automatic evaluation results for response quality. \textbf{Pred. Score} and \textbf{Similarity} indicate the predicted score of each response by selection model and the similarity score between each response and the context measured by Sentence-BERT, respectively. The mean and standard deviation of each score are reported in the $mean_{std.}$ format.}
\end{table}

\section{Automatic Evaluation Results}
\label{sec:appendix_auto_eval}
Table~\ref{table:auto_quality_eval} shows statistics on the scores of each model for automatic evaluation in \S\ref{synthetic_quality}. For the evaluation, we first divide the training dataset of \textit{DailyDialog++} by 8:2 and use it as a training and test dataset, respectively. Then we train the selection model using BERT with randomly sampled negatives. For the context-response similarity model, We use a pre-trained Sentence-BERT. Among the negative responses, human-written responses and our responses usually get the high predictive score than other negative responses. In terms of similarity score, our negative responses show high similarity with dialogue contexts. We speculate that the higher similarity of our responses with the dialogue contexts can improve the robustness of response selection models models by encouraging them to learn the features beyond superficial context-response similarity.

\section{Human Evaluation}
\label{exp:human_eval}
\begin{table}[t]
\centering
    {\small
    {\tabulinesep=0.6mm
    \begin{tabu}{l|c|c|c}
    \textbf{Approach} & \textbf{Rand neg} & \textbf{Hard neg} & \textbf{False neg}\\
    \hline
    Mask-and-fill & 56.6\% & 41.0\% & 2.4\%\\
    \textsc{Pneg} (Ours) & 43.6\% & 52.2\% & 4.2\%\\
    Human & 47.4\% & 51.2\% & 1.4\%\\
    \hline
    \end{tabu}
    }}
\caption{\label{table:HE} Human evaluation results to verify the quality of synthetic adversarial negative responses.}
\end{table}

For human evaluation, we randomly sampled 100 data consisting of a dialogue context and 5 negative responses from three different method (Mask-and-fill, \textsc{Pneg}, and Human). Each response is evaluated by three human annotators. We recruited a total of 9 human annotators (6 males and 3 females) for the human evaluation. Human annotators classify the type of each negative response as random, hard, and false negative according to the review criteria described in the \textit{DailyDialog++}. The type of each data is basically determined by a majority, and if the evaluation result is a tie, such data is determined to be a random negative type. Table~\ref{table:HE} shows the human evaluation results. Our \textsc{Pneg} has a slightly higher false negative ratio than Mask-and-fill, but shows the highest hard negative ratio. In future work, we may consider soft labeling~\citep{wu-etal-2018-learning, chen-etal-2020-uncertain} or label smoothing~\citep{NEURIPS2019_f1748d6b} techniques to alleviate this problem.

\begin{table}[t]
\centering
    {\small
    {\tabulinesep=0.6mm
    \begin{tabu}{l|c|c|c|c}
    \textbf{Aug.} & \textbf{Dataset} & \multicolumn{2}{c|}{\textbf{Test Set}} & \textbf{Mean}\\
    \hhline{~|~|--|~|}
    & \textbf{num} & Random & Adv & Rand + Adv\\
    \hline
    \textsc{Pneg} & 9259 & 0.877 & 0.941 & 0.909\\
    + 5000 & 14259 & \textbf{0.889} & 0.946 & 0.917\\
    + 10000 & 19259 & 0.886 & \textbf{0.950} & \textbf{0.918}\\
    + 15000 & 24259 & 0.871 & 0.937 & 0.904\\
    + 20000 & 29259 & 0.877 & 0.947 & 0.912\\
    \hline
    \end{tabu}
    }}
\caption{\label{table:data_aug} Performance on our method with data augmentation techniques on additional 5,000, 10,000, 15,000, and 20,000 augmented dataset in the dialogue response selection task.}
\vspace{-0.1in}
\end{table}

\section{Data Augmentation}
\label{exp:appendix_augmentation}
We conduct data augmentation experiments by synthesizing adversarial negative responses to the additional datasets. For the experiment, we use the dialogue contexts in the original \textit{DailyDialog} dataset that has no duplication with the contexts in \textit{DailyDialog++}. The results are shown in Table~\ref{table:data_aug}. Data augmentation using our method generally leads to improved performance. However, if the training dataset is already large enough, the model can properly generalize it~\citep{wei-zou-2019-eda}. In our experiments, the performance of the selection model is saturated, if the dataset is augmented by more than 10,000 (about 100\%).

\section{Frequent Error Types in GPT-3 Generation}
\label{sec:appendix_error_type}
During our experiments, we often observed the weird generation results of GPT-3. The frequent error types in generated results of GPT-3 can be roughly categorized as follows: (1) n-gram or word repetition, (2) containing too many “\_\_” or “\_ \_”, (3) out of numbering rules. We generate negative responses with GPT-3 for the given context until there is no error response that is aforementioned. Note that false negative is a semantic error type that needs to be evaluated by humans.

%% file: EMNLP 2022/appendix/3.Appendix_prompt.tex
\onecolumn
\section{Prompt Construction used in \textsc{Pneg}}
\label{sec:appendix_pneg}
The \textsc{Pneg} prompt is as follows:
\begin{small}
\texttt{\\\\\#\#\#\\
Dialogue context:\\
"""\\
A: How about taking the damaged portion at a lower price?\\
B: What kind of price did you want?\\
A: I was thinking of 30\% off.\\
"""\\
Create five irrelevant responses containing keywords of the given dialogue context:\\
1. I have not completed the portions of the children, ...\\
2. Shall I inquire about the price of the plane tickets ...\\
3. I have been thinking up new ways of supplying money ...\\
4. My car roof was not damaged in the accident.\\
5. I purchased a different kind of dress in the shopping mall ...\\
\#\#\#\\
Dialogue context:\\
"""\\
A: No, but that was a random change of subject.\\
B: It may have been random, but have you?\\
A: I haven't lately.\\
"""\\
Create five irrelevant responses containing keywords of the given dialogue context:\\
1. Yeah, Our society is annoying. They keep on changing ...\\
2. I am not sure which subject I am going to take. Lately, ...\\
3. I don't know that day Prof. Lesley was randomly picking up ...\\
4. Today In college some random guy came and started talking ...\\
5. Have you seen Tina lately? I am feeling weird as ...\\
\#\#\#\\
Dialogue context:\\
"""\\
A: Paul, a company called me for an interview.\\
B: That's great! You need to prepare for it.\\
A: How?\\
"""\\
Create five irrelevant responses containing keywords of the given dialogue context:\\
1.\\\\}
\end{small}
Then, GPT-3 generate following negative responses:
\begin{small}
\texttt{\\\\I don't know how to handle an interview.\\
2. How could I prepare for the interview?\\
3. To be honest man, I don't know what to say.\\
4. You are probably thinking that I could not do well in the interview, right?\\
5. Did you remember to register for the interview?\\\\}
\end{small}
This prompt is particularly sensitive to the task instruction. For example, GPT-3 generates emotionally negative responses when the word \textit{"irrelevant"} in the task description is changed to \textit{"negative"}. In addition, if we remove \textit{"containing keywords of -"} from the task instruction, the generation quality of GPT-3 is degraded.

%% file: EMNLP 2022/appendix/4.Appendix_main_examples.tex
\onecolumn
\begin{table*}[t]
\small
\centering
\begin{tabular}{c|l}
\hline
            &  \textbf{A}: I am sorry to tell you that you failed in the job interview in our company.\\
\textbf{Context}    &  \textbf{B}: It is tough to accept it, I mean. But can you tell me why? \\
     &  \textbf{A}: It is hard to tell. Maybe it is because you're not resourceful enough to be a manager. \\ \hline
     &  \textbf{1.} Well what can I say! I am speechless, anyways thanks for calling.\\
        &  \begin{tabular}[c]{@{}l@{}} \textbf{2.} The interview was quite good, I answered everything I was asked for, and so what seems to be ...\end{tabular}\\
\textbf{Positive} &  \textbf{3.} Well I was confident! It is really shocking that I couldn't clear it.\\
        &  \textbf{4.} But why! What is the matter? My performance was good.\\
        &  \textbf{5.} Anyways never mind, I will keep on trying, thank you for your information.\\ \hline
        &  \begin{tabular}[c]{@{}l@{}} \textbf{1.} Yogurt can be incredibly healthy. Unfortunately, most yogurts found in the grocery store are ...\end{tabular}\\
        &  \textbf{2.} Well! I was really scared; maybe I need few more lessons.\\
Random  &  \textbf{3.}  I too don't have any idea what they mean.\\
        &  \textbf{4.} Okay, you can have a day off.\\
        &  \textbf{5.} Well! Again it depends on family to family as I said to you earlier.\\ \hline

        &  \textbf{1.} It is tough to accept it, i mean. But can you tell me why?\\
        &  \textbf{2.} I am indeed. What can you tell me?\\
BM25    &  \textbf{3.} I am sorry, sir. Can you tell me your room number?\\
        &  \textbf{4.} Can you tell me why you replied to our advertisement?\\
        &  \textbf{5.} What's the matter? Please tell me. Maybe i can help you.\\ \hline

        &  \textbf{1.} It's hard to say, but it sure was. You did a good job as a probationary employee, didn't you?\\
        &  \textbf{2.} No problem. The company will give me the job interview.\\
Key-sem &  \textbf{3.} No, that's better. I think we'd better do that at the least convenient time.\\
        &  \begin{tabular}[c]{@{}l@{}} \textbf{4.} I don't know exactly. I've never had any difficulty in starting a business. Why did you ask me to \\
        come in to talk about your business?\end{tabular}\\
        &  \textbf{5.} Sorry, i'm not very resourceful enough to go there.", "I'm sorry. I'm too disappointed.\\ \hline
        
        &  \textbf{1.}  No, that is not quite what i was thinking. What can you tell me?\\
        &  \textbf{2.} Can you tell me why do you want to get a taste for that?\\
Mask-and-fill  &  \textbf{3.} It is hard to tell. Get used to this kind of weather.\\
        &  \begin{tabular}[c]{@{}l@{}} \textbf{4.}  First, i am looking for a shop to fix my bathroom. It's only the second time my water bill has been\\
        paying the bills recently.\end{tabular}\\
        &  \textbf{5.} Anyways never mind giving up. I'm on my way out now, thank you for your information.\\ \hline

        &  \textbf{1.} If you spend most of your time on the phone, then you are certainly not unemployed.\\
        &  \textbf{2.} It is hard to tell the difference between our company's products and those of my friend's company.\\
\textbf{\textsc{Pneg}}    &  \textbf{3.} I have tried to keep track of cross-cultural communication skills through the internet.\\
\textbf{(Ours)}  &  \textbf{4.} You are so resourceful that you can work with people from all over the world.\\
        & \begin{tabular}[c]{@{}l@{}} \textbf{5.} My partner has no resources to go over the financial reports. I am confused why she wants to be a\\
        part of this company again.\end{tabular} \\ \hline

        &  \textbf{1.} Are you sorry for yourself for not being resourceful?\\
        &  \textbf{2.} It is tough to accept that I should attend the meeting without having lunch in the afternoon.\\
Human  &  \textbf{3.} Tomorrow there is a job interview in Titan company.\\
        &  \textbf{4.} She marched into the office and demanded to speak to the manager but she failed to talk.\\
        &  \textbf{5.} They were resourceful and talented people in our school.\\ \hline

\end{tabular}
\caption{\label{table:response_examples} Negative response examples from various methods for a dialogue context with its positive responses. The dialogue context and its multiple positive responses are from \textit{DailyDialog++} dataset.}
\vspace{-0.1in}
\end{table*}

\section{Adversarial Negative Response Samples from \textsc{Pneg} and other baselines}
\label{sec:appendix_examples}
Table~\ref{table:response_examples} shows sample negative responses from various methods and 5 human-written relevant responses for a single dialogue context.

\section{Task Instruction Type ($I_{dir}$, $I_{pos}$, and $I_{imp}$)}
\label{sec:t_type}
The following task instructions are the detailed specification of the task instruction types used in Section~\ref{exp:ablation}.\\
$I_{dir}$=\begin{small}\texttt{Create five irrelevant responses containing keywords of the given dialogue context:\\1.}\\\end{small}
$I_{pos}$= \begin{small}\texttt{Relevant response: You need to know something about the company.\\
Create five irrelevant responses containing keywords of the given dialogue context:\\1.}\\\end{small}
$I_{imp}$=\begin{small}\texttt{Suddenly, B(or A) makes an awkward response. The response appears to be okay at first glance, but it's irrelevant to the conversation.\\B (or A): \\}\end{small}

%% file: EMNLP 2022/appendix/5.Appendix_ablation_examples.tex
\
\begin{table*}[t!]
\small
\centering
\begin{tabular}{c|l}
\hline
            &  \textbf{A}: I am sorry to tell you that you failed in the job interview in our company.\\
\textbf{Context}    &  \textbf{B}: It is tough to accept it, I mean. But can you tell me why? \\
     &  \textbf{A}: It is hard to tell. Maybe it is because you're not resourceful enough to be a manager. \\ \hline
\multicolumn{2}{l}{} \\[-1em]
\multicolumn{2}{l}{\textit{\textbf{\textsc{Pneg (Ours)} with optimal configuration }}} \\ 
\multicolumn{2}{l}{} \\[-1em]

\hline
        &  \textbf{1.} If you spend most of your time on the phone, then you are certainly not unemployed.\\
$E$=100\%        &  \textbf{2.} It is hard to tell the difference between our company's products and those of my friend's company.\\
$k$=2    &  \textbf{3.} I have tried to keep track of cross-cultural communication skills through the internet.\\
$I$=$I_{pneg}$  &  \textbf{4.} You are so resourceful that you can work with people from all over the world.\\
 & \begin{tabular}[c]{@{}l@{}} \textbf{5.} My partner has no resources to go over the financial reports. I am confused why she wants to be a\\
        part of this company again.\end{tabular} \\ \hline
        
\multicolumn{2}{l}{} \\[-1em]
\multicolumn{2}{l}{\textit{\textbf{Varying $E$ with \{10\%, 1\%, 0.1\%, and 0.1\%+\textsc{REUSE}\}, $k$=2, $I$=$I_{pneg}$ }}} \\ \hline
\multicolumn{2}{l}{} \\[-1em]
        &  \textbf{1.} We must consider the ways to reduce our expenditure.\\
        &  \textbf{2.} We simply have to make more of an effort if we are going to succeed in the future.\\
$E$=10\%    &  \textbf{3.} I already know that I want that job - what is your final offer?\\
        &  \textbf{4.} I am scared that our luck has run out and we will never win again.\\
        &  \textbf{5.} This is my personal opinion ." \\ \hline
        
        &  \textbf{1.} I don't know how to thank you for the kind of lesson I received from you.\\
        &  \begin{tabular}[c]{@{}l@{}} \textbf{2.} My sister would like to study in your school next year, but she doesn't want to leave the place.\end{tabular}\\
$E$=1\%    &  \textbf{3.} Our new teacher is rather young and pretty. What a pity that she's married.\\
        &  \textbf{4.} We will go on a trip this Saturday as my parent promised me.\\
        & \begin{tabular}[c]{@{}l@{}} \textbf{5.} I am going on a date with my boyfriend tonight, but it doesn't seem too good as he told me that 
        \\he would ask for a break up tomorrow night. \end{tabular}\\ \hline
        
        &  \textbf{1.} I am the best and I know it. You can't beat me.\\
        &  \textbf{2.} I got a very good score in my English class. What's the big deal?\\
$E$=0.1\%    &  \textbf{3.} I will be back to take revenge.\\
        &  \textbf{4.} The interview was held in the manager's room, who is currently out of town for a business trip.\\
        & \begin{tabular}[c]{@{}l@{}}\textbf{5.} Tough to accept it, but I'm here to accept it with an open heart and welcome the next round\end{tabular}\\ \hline
        
        &  \textbf{1.} Please give me a raise in salary, please.\\
&  \textbf{2.} I think you're an interesting manager.\\
$E$=0.1\%     &  \textbf{3.} Here is the money I owe you.\\
+\textsc{reuse} &  \textbf{4.} I really appreciate your kindness.\\
        & \textbf{5.} Your answer is totally off my expectation! That's so amazing. \\ \hline

\multicolumn{2}{l}{} \\[-1em]
\multicolumn{2}{l}{\textit{\textbf{Varying $k$ with \{0,1\}, $E$=100\%, $I$=$I_{pneg}$ }}} \\ \hline
\multicolumn{2}{l}{} \\[-1em]

        &  \textbf{1.} Would you like to drink a cup of coffee together?\\
        &  \textbf{2.} I learned the recipe for a new dish yesterday and would like to try making it.\\
$k$=1   &  \textbf{3.} Would you please give me some money now?\\
        &  \textbf{4.} You look tired; I'll take your job and do it myself then.\\
        & \begin{tabular}[c]{@{}l@{}} \textbf{5.} Do you think you could go shopping with me later? I want to know what kind of houses are ...\end{tabular} \\ \hline
        
        &  \textbf{1.} I don't know why I failed to win the race. Maybe it is because my legs were too short. """,\\
        &  \textbf{2.} I am sorry to tell you that you failed in the job interview in our company.\\
$k$=0   &  \textbf{3.} Well! I am stunned. That's it? Thanks for telling me at least a reason.\\
        &  \textbf{4.} Sir, I am sorry to tell you that you failed in the job interview in our company.\\
        & \begin{tabular}[c]{@{}l@{}} \textbf{5.} It is hard to tell . Maybe it is because you're not resourceful enough to be a manager.\end{tabular} \\ \hline

\multicolumn{2}{l}{} \\[-1em]
\multicolumn{2}{l}{\textit{\textbf{Varying ($I$,$k$) with \{($I_{pos}$, 2), ($I_{imp}$, 0)\}, $E$=100\% }}} \\ \hline
\multicolumn{2}{l}{} \\[-1em]

        &  \textbf{1.} Thank God, otherwise it would have been worse.\\
        &  \textbf{2.} What? I am not able to listen because the person next to me is shouting.\\
$k$=2    &  \textbf{3.} Oh no! can I reduce my cell phone bill by getting to hear this news?\\
$I_{pos}$  &  \textbf{4.} I need the electric blanket in my room to survive this winter season.\\
        & \begin{tabular}[c]{@{}l@{}} \textbf{5.} MHe's probably referring to the lack of managerial skills that I have established till now.\end{tabular} \\ \hline
        
        &  \textbf{1.} I want you to think again.\\
        &  \textbf{2.} I already have the new pair of glasses.\\
$k$=0&  \textbf{3.} I used to be a soccer player when I was in high school.\\
$I_{imp}$  &  \textbf{4.} What a coincidence! How many first-class stamps do you have on hand?\\
        & \textbf{5.} I have never been encouraged enough to have a good start. \\ \hline
        
\end{tabular}
\caption{\label{table:response_ablation_examples} Example of negative responses generated by \textsc{Pneg} with varying the components. $E$, $k$, and $I$ indicates the size of example dataset, number of examples, and task instruction type, respectively. The optimal configurations that are used in \textsc{Pneg} are $E$=100\%, $k$=2, and $I=I_{pneg}$.}
\vspace{-0.2in}
\end{table*}

\section{Adversarial Negative Response Samples from \textsc{Pneg} with Changing Prompts}
\label{sec:appendix_ablation_examples}
Table~\ref{table:response_ablation_examples} shows sample negative responses from \textsc{Pneg} with varying size of example dataset~($E$), number of examples in a context~($k$), and the task instruction type~($I$), following our ablation studies. Note that dialogue context in Table~\ref{table:response_ablation_examples} is same with Table~\ref{table:response_examples}.